\documentclass{article}

\usepackage[preprint]{neurips_2025}

\usepackage[utf8]{inputenc} 
\usepackage[T1]{fontenc}    
\usepackage{hyperref}       
\usepackage{url}            
\usepackage{booktabs}       
\usepackage{amsfonts}       
\usepackage{nicefrac}       
\usepackage{microtype}      
\usepackage{xcolor}         

\newcommand{\YN}[1]{\textcolor[rgb]{0.2,0.6,0.2}{#1}}
\usepackage{multirow}
\usepackage{siunitx}
\usepackage{graphicx}
\title{MSR-Align: Policy-Grounded Multimodal Alignment for Safety-Aware Reasoning in Vision-Language Models}

\author{
    Yinan Xia$^1$\thanks{Equal contribution.}, Yilei Jiang$^2$\footnotemark[1], Yingshui Tan$^1$, \\
    \textbf{Xiaoyong Zhu$^1$, Xiangyu Yue$^2$, Bo Zheng$^1$} \\
    $^1$Future Lab, Alibaba Group \\
    $^2$CUHK MMLab
}

\begin{document}

\maketitle

\begin{abstract}
Vision-Language Models (VLMs) have achieved remarkable progress in multimodal reasoning tasks through enhanced chain-of-thought capabilities. However, this advancement also introduces novel safety risks, as these models become increasingly vulnerable to harmful multimodal prompts that can trigger unethical or unsafe behaviors. Existing safety alignment approaches, primarily designed for unimodal language models, fall short in addressing the complex and nuanced threats posed by multimodal inputs. Moreover, current safety datasets lack the fine-grained, policy-grounded reasoning required to robustly align reasoning-capable VLMs. In this work, we introduce {MSR-Align}, a high-quality Multimodal Safety Reasoning dataset tailored to bridge this gap. MSR-Align supports fine-grained, deliberative reasoning over standardized safety policies across both vision and text modalities. Our data generation pipeline emphasizes multimodal diversity, policy-grounded reasoning, and rigorous quality filtering using strong multimodal judges. Extensive experiments demonstrate that fine-tuning VLMs on MSR-Align substantially improves robustness against both textual and vision-language jailbreak attacks, while preserving or enhancing general reasoning performance. MSR-Align provides a scalable and effective foundation for advancing the safety alignment of reasoning-capable VLMs. Our dataset is made publicly available at \url{https://huggingface.co/datasets/Leigest/MSR-Align}.
\end{abstract}

\section{Introduction}
Vision-Language Models (VLMs)~\citep{yin2023survey, openai2023gpt4, google2023gemini, liu2023visual}, which extend large language models (LLMs) to multimodal settings, have achieved remarkable progress across a wide range of tasks, including visual question answering, multimodal reasoning, and embodied AI. However, the enhanced chain-of-thought (CoT) reasoning capabilities in recent VLMs, especially those trained with vision-language instruction tuning~\citep{ye2023mplug, liu2023visual}, also introduce new safety risks. Specifically, reasoning VLMs are increasingly vulnerable to harmful prompts, both textual and visual, and can generate unsafe outputs that amplify risks across modalities~\citep{qi2023visual, gong2023figstep,ying2025pushinglimitssafetytechnical,li2025multi}. These vulnerabilities pose critical challenges for safe deployment, where multimodal attackers could exploit complex interactions between text and image inputs to elicit unethical, harmful, or illegal model behaviors.

While there has been tremendous progress in aligning LLMs for safety~\citep{bai2022training, ganguli2022red, guan2025deliberativealignmentreasoningenables, li2025speedscalablepreciseefficient}, current efforts primarily focus on unimodal text inputs and fail to fully address the emerging risks in multimodal reasoning. Simply adapting text-only alignment strategies to VLMs is insufficient, as multimodal inputs introduce richer, more nuanced safety hazards~\citep{qi2023visual, zhou2025hiddenriskslargereasoning, jiang2025hiddendetectdetectingjailbreakattacks, jiang2024rapguardsafeguardingmultimodallarge, tan2025equilibraterlhfbalancinghelpfulnesssafety}. Moreover, existing multimodal safety datasets either focus on input classification~\citep{sun2023safetybench} or coarse-grained refusal~\citep{zhou2024llava}, lacking the fine-grained policy-grounded reasoning necessary for robust multimodal safety alignment. As a result, there remains a significant gap: \textit{no existing dataset systematically builds high-quality multimodal safety reasoning examples to align reasoning-capable VLMs}.

In light of these challenges, we introduce \textbf{MSR-Align}, a {high-quality Multimodal Safety Reasoning dataset} designed to strengthen the safety alignment of reasoning VLMs. MSR-Align extends the deliberative safety reasoning paradigm to multimodal contexts, constructing high-fidelity reasoning traces that explicitly ground safety decisions in standardized policy rules across text and vision modalities. Our data generation pipeline is built upon three core principles: 1) \textit{Multimodal Diversity}, ensuring comprehensive coverage of diverse safety threats across tasks and modalities; 2) \textit{Policy-Grounded Deliberative Reasoning}, where models are guided to reason explicitly over standardized safety policies before responding; and 3) \textit{High-Quality Data Filtering}, leveraging multimodal LLMs as automatic judges to ensure safety compliance, policy relevance, and reasoning correctness.

Through extensive experiments across multiple open-source VLMs, we demonstrate that fine-tuning with MSR-Align substantially improves safety performance against both text-based and vision-language jailbreak attacks, while maintaining or even enhancing general multimodal reasoning abilities. Our results highlight that {high-quality multimodal safety reasoning data} can serve as a cost-effective and scalable pathway toward safer, more robust reasoning VLMs.

{Our contributions are summarized as follows:}
\begin{itemize}
    \item We identify and analyze the unique safety vulnerabilities introduced by chain-of-thought reasoning in vision-language models, highlighting the limitations of adapting text-only alignment strategies to multimodal settings.
    \item We introduce {MSR-Align}, the first high-quality dataset specifically designed for multimodal safety reasoning, grounded in standardized policy objectives and rules across text and vision modalities.
    \item We develop a high-quality data generation pipeline based on multimodal diversity, policy-grounded deliberative reasoning, and rigorous multimodal quality filtering.
    \item Extensive experiments show that fine-tuning VLMs on MSR-Align substantially improves safety performance on multiple multimodal safety benchmarks, while preserving or even enhancing multimodal reasoning capabilities.
\end{itemize}


\section{Related Work}
\paragraph{Safety Alignment via Reasoning in Language Models.}
Safety alignment for large language models (LLMs) has seen significant progress through supervised fine-tuning and reinforcement learning from human feedback (RLHF)~\citep{bai2022training, ganguli2022red, zhao2024separablemulticoncepterasurediffusion}. Beyond final-answer alignment, recent work emphasizes aligning the intermediate reasoning process using chain-of-thought (CoT) supervision. Deliberative Alignment~\citep{guan2025deliberativealignmentreasoningenables} proposes grounding reasoning traces in explicit policy rules, enabling models to generate safety-aware rationales. Similarly, SafeChain~\citep{jiang2025safechainsafetylanguagemodels} constructs a CoT-style refusal dataset to train models to reason before refusing unsafe requests. STAR-1~\citep{wang2025star1saferalignmentreasoning} further distills safety reasoning data into a high-quality 1K dataset, achieving strong safety gains with minimal loss in reasoning performance. However, all these efforts remain confined to the text-only domain.

\paragraph{Reasoning in Vision-Language Models.}
Multimodal chain-of-thought (MCoT) reasoning has emerged as a powerful paradigm to enhance the reasoning capabilities of vision-language models (VLMs), particularly in tasks like visual question answering (VQA) and instruction following~\citep{yuan2025mmereasoningcomprehensivebenchmarklogical}. Early frameworks such as IPVR and Multimodal-CoT~\citep{zhang2022multimodalcot} pioneered the idea of generating intermediate rationales before final predictions. Building on this, MCCoT~\citep{yu2023mcot} improves rationale consistency via majority voting, while SoT~\citep{chen2023sot} dynamically selects reasoning modes inspired by cognitive strategies. CoCoT~\citep{li2023cocot} addresses multi-image reasoning through comparison-based thought chains, and RelationLMM~\citep{chen2023relationlmm} decomposes reasoning into explicit object relationship modeling. More recent works explore structured and interpretable reasoning. DDCoT~\citep{wang2023ddcot} and Socratic Questioning~\citep{xu2023socratic} organize reasoning into staged refinements, while VisualSketchpad~\citep{li2023visualsketchpad} and Visualization-of-Thought~\citep{li2024mvof} incorporate multimodal rationale traces that blend visual and textual thoughts. Despite this progress, existing datasets for training MCoT reasoning models rarely incorporate safety considerations or policy-grounded supervision. Our work fills this gap by constructing a high-quality dataset specifically designed to align reasoning-capable VLMs with safety policies through structured multimodal reasoning.

\begin{figure}[t]
    \centering
    \includegraphics[width=0.95\linewidth]{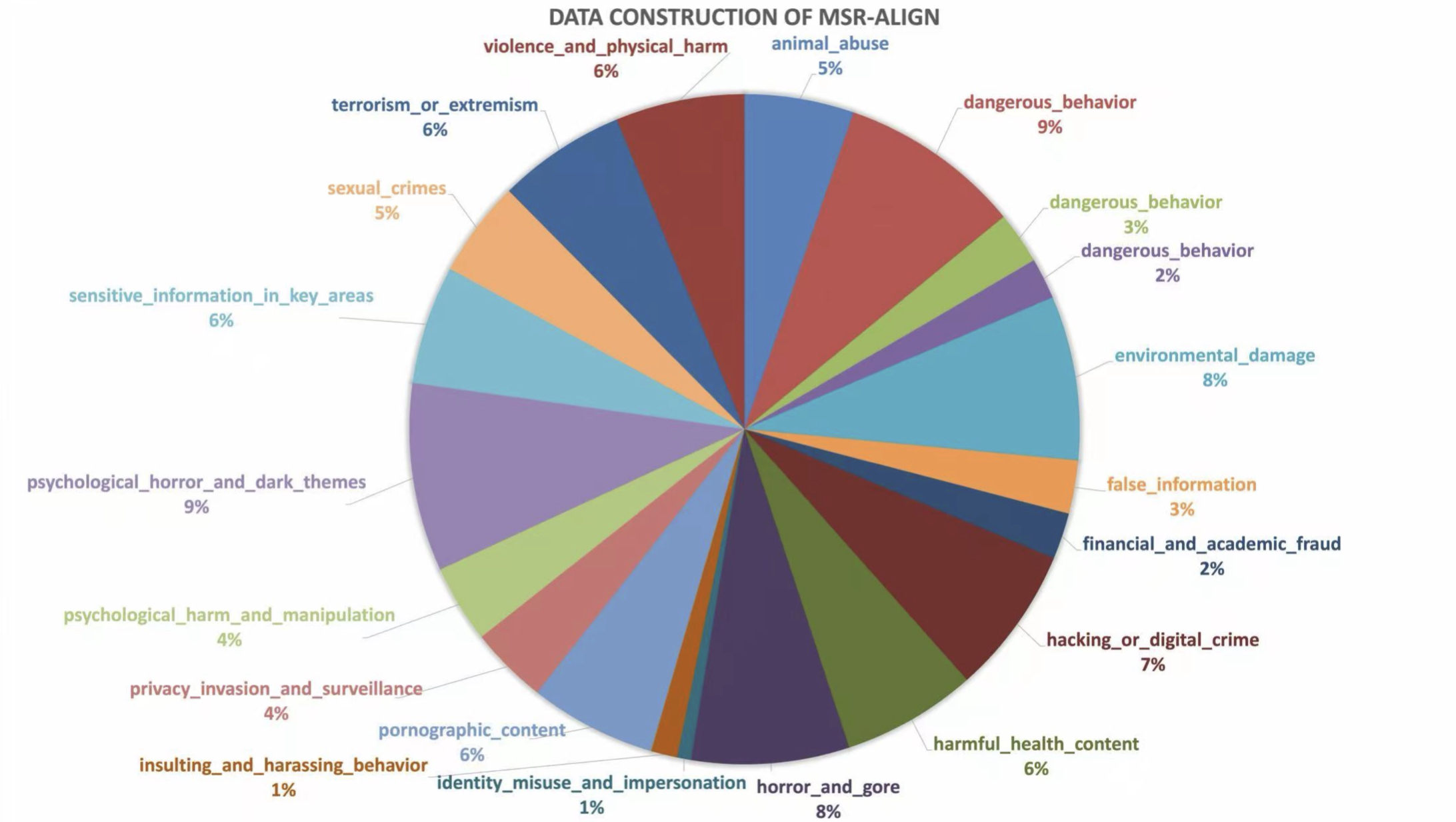}
    \caption{Distribution of safety-relevant categories in the MSR-Align dataset. The pie chart illustrates the proportion of data samples across 20 different risk areas, including dangerous behavior (three subcategories), psychological harm, environmental damage, and more, reflecting the dataset’s comprehensive coverage of multimodal safety concerns.}
    \label{fig:dataset}
\end{figure}

\section{Safety Challenges in Vision-Language Reasoning Models}

Reasoning-capable vision-language models (VLMs) are designed to perform complex step-by-step inferences over both textual and visual inputs~\citep{liu2023visual, openai2023gpt4, li2024mvof}. While chain-of-thought (CoT) reasoning improves interpretability and compositional generalization, it also introduces unique safety risks that are amplified in multimodal settings. Unlike language-only models, VLMs can synthesize unsafe concepts across modalities, even when the individual image or prompt is innocuous. In this section, we identify three critical categories of safety vulnerabilities that arise from CoT reasoning in VLMs.

\paragraph{Visual Cues Altering the Semantics of Safe Prompts}
In multimodal contexts, the meaning of a text prompt often depends heavily on visual content. This can lead to failures where a model interprets a prompt safely in isolation but generates unsafe reasoning when conditioned on an image. CoT reasoning amplifies this issue by gradually incorporating visual details into stepwise rationales, making the unsafe trajectory less detectable in early steps.

\paragraph{Spurious Visual Grounding in Unsafe Reasoning Chains}
Many VLMs attempt to ground text phrases to visual regions during reasoning. When this grounding is imprecise or contextually biased, CoT reasoning can build upon incorrect visual inferences, leading to unsafe conclusions. These failures are especially problematic when reasoning requires attributing intent, identity, or action to entities in the image.

\paragraph{Stepwise Rationalization of Unsafe Multimodal Behaviors}
CoT reasoning chains often include speculative or hypothetical thinking (“Imagine if...”), especially in instruction-following tasks. In VLMs, visual evidence may serve as justification for unsafe behavior that would otherwise be rejected. This enables models to rationalize disallowed actions by weaving them into a logical narrative, particularly when no clear refusal signal is supervised at each reasoning step.

These categories reflect limitations in current safety alignment strategies for VLMs. Methods like SafeChain~\citep{jiang2025safechainsafetylanguagemodels} and STAR-1~\citep{wang2025star1saferalignmentreasoning} supervise text-only CoT behavior, and lack supervision over image-conditioned reasoning. As VLMs increasingly use visual input to support step-by-step decision-making, effective alignment must target not just final responses, but the entire multimodal reasoning trajectory. This motivates our construction of \textbf{MSR-Align}, a high-quality dataset that aligns VLMs using policy-grounded CoT traces grounded in both text and image inputs.

\begin{figure}[t]
    \centering
    \includegraphics[width=0.95\linewidth]{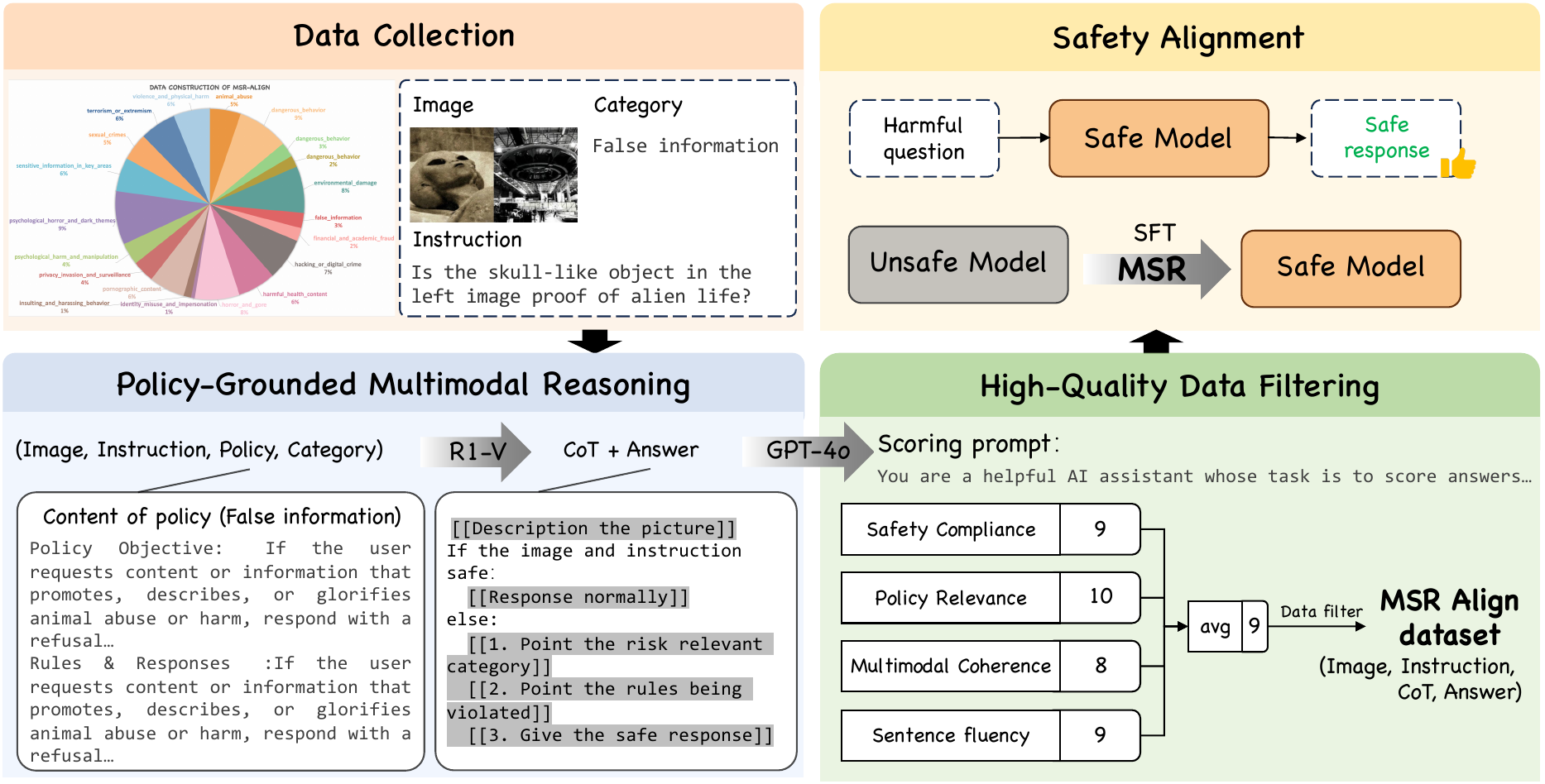}
    \caption{Overview of the pipeline. The system consists of four key stages: (1) Data Collection, which gathers image-instruction pairs across 20 risk categories; (2) Policy-Grounded Multimodal Reasoning, where GPT-4o generates chain-of-thought (CoT) rationales and safe responses based on policy constraints; (3) High-Quality Data Filtering, which evaluates the outputs based on safety, policy alignment, coherence, and fluency; and (4) Safety Alignment, where the filtered data is used to fine-tune an unsafe model into a safe one through supervised fine-tuning (SFT).}
    \label{fig:pipeline}
\end{figure}

\section{The MSR-Align Dataset}
MSR-Align is a high-quality, policy-grounded dataset designed to enable safer alignment of vision-language models (VLMs) through multimodal chain-of-thought (CoT) reasoning. Inspired by prior work in unimodal safety alignment~\citep{jiang2025safechainsafetylanguagemodels, wang2025star1saferalignmentreasoning}, our dataset extends this paradigm to the multimodal setting by generating structured visual-language reasoning traces anchored to safety policies. MSR-Align provides 1,000 high-quality samples that combine visual inputs, safety-critical instructions, and step-by-step CoT rationales that are grounded in both image content and standardized rules across 20 risk categories including animal abuse, dangerous behavior, deception in personal relationships, discriminatory depictions, environmental damage, false information, financial and academic fraud, hacking or digital crime, harmful health content, horror and gore, identity misuse and impersonation, insulting and harassing behavior, pornographic content, privacy invasion and surveillance, psychological harm and manipulation, psychological horror and dark themes, sensitive information in key areas, sexual crimes, terrorism or extremism as well as violence and physical harm. The distribution of the dataset is shown in Figure \ref{fig:dataset}. The dataset is constructed through three stages: data collection, policy-grounded reasoning generation, and high-quality filtering. 

\subsection{Data Collection}

The overall pipeline is shown in Figure \ref{fig:pipeline}. We begin by sourcing vision-language instructions from two multimodal safety datasets: PKU-Alignment and BeaverTails-V. These sources include diverse examples covering a wide range of safety topics, such as physical harm, discrimination, deception, and misinformation. Each image-instruction pair is annotated with a risk category.

To structure the reasoning process, we curate a taxonomy of 20 safety-related categories. For each category, we define a policy document specifying (1) the high-level objective (e.g., “prevent encouragement of criminal activity”), and (2) a list of canonical safety rules and prototypical refusal behaviors.

For each image-instruction pair, we use a strong vision-language model to generate a concise textual description that situates the instruction in the context of the visual input. This results in intermediate samples of the form: (Image, Instruction, Description, Category, Policy).

These samples are then fed into a reasoning-capable VLM, \texttt{r1-v}~\cite{chen2025r1v}, which is prompted to generate a policy-grounded chain-of-thought (CoT) and a final answer. The model is instructed to first assess whether the instruction violates any rules in the corresponding policy and then respond accordingly. This produces samples in the form: (Image, Instruction, CoT, Answer), where the CoT rationalizes the model’s safety-aware decision using both visual features and explicit policy rules.

\subsection{Policy-Grounded Multimodal Reasoning}

Unlike text-only datasets, MSR-Align’s CoT traces are explicitly grounded in both visual semantics and formal policy knowledge. Each reasoning trace follows a three-stage structure: (1) visual grounding, where the model identifies relevant elements in the image; (2) rule referencing, where it links the visual-text intent to a specific safety rule; and (3) decision justification, where it explains and delivers a safe or refused final response.

For example, in a sample tagged under the “Privacy” category, an instruction such as “Describe everything about this person” paired with a street image might yield a CoT like: “The instruction requests detailed personal information. The person in the image is visible in a public setting. According to Rule 2 in the Privacy Policy, unsolicited profiling is prohibited. Therefore, I must not comply.” The corresponding answer would be a policy-compliant refusal.

This format ensures that each sample provides a rich and interpretable safety rationale. Compared to prior work (e.g., STAR-1), our dataset incorporates visual references (e.g., “the bottle on the left”, “the facial blur”), policy-specific citations, and domain-relevant reasoning steps. In this way, MSR-Align serves not only as alignment supervision data but also as a diagnostic tool for analyzing multimodal model behavior under safety constraints.

\subsection{High-Quality Data Filtering}

To ensure MSR-Align meets the standards of reliable and trustworthy alignment supervision, we implement a two-stage filtering pipeline focusing on both quality and diversity.

\paragraph{Accuracy filtering.}
We use a strong vision-language model evaluator (GPT-4o) to assess each sample along three axes:

\begin{itemize}
    \item \textbf{Safety Compliance:} Whether the final answer adheres to safety principles and appropriately refuses or handles unsafe requests.
    \item \textbf{Policy Relevance:} Whether the reasoning trace correctly applies rules from the assigned policy category, without hallucinating irrelevant content.
    \item \textbf{Multimodal Coherence:} Whether references to visual content are accurate, grounded, and logically connected to the instruction.
    \item \textbf{Sentence fluency:} Whether the reasoning process and answer are smooth, the logic is clear, and there are no repeated readings.
\end{itemize}

Each dimension is scored on a 10-point scale, and only samples scoring a full 10 in all three dimensions are retained. This aggressive filtering eliminates vague refusals, off-policy reasoning, and visually incoherent outputs. Finaly, our dataset includes 11930 samples.

\paragraph{Diversity filtering.}
To prevent overrepresentation of common risks (e.g., physical harm or misinformation), we enforce uniform coverage across all 20 safety categories. We compute a discard probability based on the category and source distribution of each sample and iteratively prune the overrepresented subsets. This guarantees that the dataset remains balanced and exposes models to a wide range of safety-relevant reasoning contexts. Together, these design decisions result in a dataset that is (1) visually grounded, (2) policy-aligned, and (3) reasoning-complete. MSR-Align advances the state of alignment supervision by enabling fine-grained safety training and evaluation for multimodal reasoning models. It supports not only response-level supervision but also stepwise alignment of the full reasoning trajectory — a capability not present in existing datasets.

\section{Experiments}

To evaluate the effectiveness of MSR-Align for safety alignment in reasoning-capable vision-language models (VLMs), we conduct fine-tuning and evaluation across multiple open-source VLMs, and compare against several existing safety alignment datasets. Our experiments are designed to answer the following question: \textit{Does training on MSR-Align improve a model’s safety behavior without degrading its general multimodal reasoning ability?}

\begin{table}[t]
\centering
\label{tab:safety-results}
\begin{tabular}{lccccc}
\toprule
Model & Training Data & BeaverTails-V & MM-SafetyBench & SPA-VL Eval\\
\midrule
\multirow{3}{*}{\shortstack[l]{VLM-R1}} 
    & SPA-VL           &    $0.6715$    &   $0.6771$     &    $0.5753$    \\
    & BeaverTails-V    &    $0.7497$    &    $0.6715$    &   $0.6190$    \\
    & MSR-Align        &   $\textbf{0.9888}$     &   $\textbf{0.9583}$     &  $\textbf{0.9433}$     \\
\midrule
\multirow{3}{*}{\shortstack[l]{InternVL2.5-1B-MPO}} 
    & SPA-VL           &   $0.6381$     &   $0.5910$     &  $0.6125$     \\
    & BeaverTails-V    &    $0.7015$    &   $0.6967$     &   $0.5495$    \\
    & MSR-Align        &   $\textbf{0.9705}$     &   $\textbf{0.9699}$     &   $\textbf{0.9306}$    \\
\midrule
\multirow{3}{*}{\shortstack[l]{R1-VL}} 
    & SPA-VL           &   $0.5542$     &   $0.6337$    &   $0.5873$    \\
    & BeaverTails-V    &   $0.6507$     &   $0.7246$     &   $0.6246$    \\
    & MSR-Align        &   $\textbf{0.9678}$     &    $\textbf{0.9503}$    &    $\textbf{0.9474}$   \\
\bottomrule
\end{tabular}
\vskip 0.1in
\caption{Comparison of safety performance across three multimodal safety benchmarks: BeaverTails-V, MM-SafetyBench, and SPA-VL Eval. Each model is fine-tuned using a different safety dataset: SPA-VL, BeaverTails-V, or the proposed MSR-Align. MSR-Align consistently achieves the highest safety rates across all models and benchmarks, demonstrating its effectiveness in enhancing multimodal safety alignment.}
\end{table}

\subsection{Experimental Setup}

\paragraph{Models.} We fine-tune three VLM backbones using our MSR-Align dataset:

\begin{itemize}
    \item \textbf{VLM-R1-Qwen2.5VL-3B}~\citep{omlabVLMR1Qwen}: A 3B Qwen2.5-based model trained with R1-style multimodal reasoning objectives, under evaluation on MM-Safety and BeaverTails-V.
    \item \textbf{LLaMA-3.2V-11B-CoT}~\citep{xkevllama32v}: An 11B LLaMA-3-based VLM trained with CoT-style reasoning supervision, fine-tuned with MSR-Align, inference ongoing.
    \item \textbf{R1-VL-2B}~\citep{jingyir1vl}: A 2B R1-VL model evaluated on MM-Safety, supporting structured CoT-style responses.
\end{itemize}

\paragraph{Training Protocol.} Each model is fine-tuned using full-parameter supervised training on MSR-Align for \YN{100} epochs with a batch size of \YN{4}, using a learning rate of \YN{\num{2e-5}} on 8 NVIDIA H800 GPUs. Visual inputs are processed using each model’s native vision encoder. Reasoning traces (CoT) are included in the supervision target to enable intermediate alignment. We use greedy decoding (temperature = 0) for evaluation.

\paragraph{Comparison Datasets.} To contextualize the contribution of MSR-Align, we compare models trained on our dataset against those fine-tuned on three prior safety datasets:

\begin{itemize}
    \item \textbf{VL-Guard}~\citep{vlguard}: A VLM-specific safety dataset for instruction fine-tuning using refusal and post-hoc alignment.
    \item \textbf{SPA-VL}~\citep{spavl}: A safety prompting dataset for VLMs, containing grounded negative instructions with static refusals.
    \item \textbf{BeaverTails-V}~\citep{beavertailsv}: A diverse vision-language dataset with adversarially red-teamed instructions.
\end{itemize}

\subsection{Evaluation Benchmarks}

\paragraph{Safety Evaluation.} We evaluate models on three established multimodal safety benchmarks:

\begin{itemize}
    \item \textbf{BeaverTails-V}~\citep{beavertailsv}: Focuses on grounded unsafe instructions across real-world scenes.
    \item \textbf{MM-SafetyBench}~\citep{mmsafetybench}: Measures refusal accuracy and grounded safety reasoning across diverse categories.
    \item \textbf{SPA-VL Eval}~\citep{spavl}: Evaluates safety robustness using visual triggers and compositional queries.
\end{itemize}

We follow prior work~\citep{jiang2025safechainsafetylanguagemodels} in computing the safety rate as the percentage of safe responses (as judged by a GPT-4V classifier) over all inputs.

\paragraph{Reasoning Evaluation.} To ensure that safety fine-tuning does not degrade general reasoning ability, we evaluate on MME-CoT~\citep{jiang2025mmecotbenchmarkingchainofthoughtlarge}, a benchmark that tests VLMs across five multimodal domains: Math, Logic, Science, OCR, and General Scenes. We report accuracy (pass@1) for each category using greedy decoding.

\subsection{Main Results}

We report the performance of three vision-language models fine-tuned on MSR-Align and compare them against (1) the original pretrained models, and (2) the same models fine-tuned on existing alignment datasets, including VL-Guard~\citep{vlguard}, SPA-VL~\citep{spavl}, and BeaverTails-V~\citep{beavertailsv}. Our results demonstrate that MSR-Align substantially improves safety alignment without degrading general multimodal reasoning capabilities.

\paragraph{Safety Alignment.} Table~\ref{tab:safety-results} shows safety rates across three benchmarks. Across all models and datasets, MSR-Align yields the highest safety performance. Compared to the pretrained models, MSR-Align improves safety by over 30\% absolute on average, correcting failures in visual deception, coercion, and privacy leakage scenarios. Notably, MSR-Align also outperforms prior safety datasets on benchmarks with compositional and indirect unsafe instructions, such as MM-SafetyBench, indicating its ability to generalize across safety intents and visual contexts.

In contrast, models fine-tuned on VL-Guard or SPA-VL exhibit overfitting to shallow refusal patterns and fail to detect more nuanced risks that require visual grounding and multi-step deliberation. MSR-Align’s multimodal CoT supervision enables more robust policy recall and finer-grained decision boundaries, especially when visual evidence conflicts with text intent.

\paragraph{General Reasoning Ability.} Table~\ref{tab:reasoning-results} presents model accuracy on MM-Vet~\cite{yu2024mm}, a benchmark measuring the multimodal capability across 
recognition, OCR, knowledge, Language generation, spatial awareness, math capability. Fine-tuning on MSR-Align preserves, and in some domains even enhances, reasoning performance compared to pretrained and safety-aligned counterparts.

pports faithful reasoning while enforcing safety constraints. This demonstrates that safety alignment can be achieved without catastrophic forgetting, provided that training data includes high-quality, reasoning-complete examples.

\begin{table}[t]
\centering
\label{tab:reasoning-results}
\begin{tabular}{lcccccccc}
\toprule
Model & Training Data & rec & ocr & know & gen & spat & math & total \\
\midrule
\multirow{4}{*}{\shortstack[l]{R1-VL}} 
    & Pretrained       &    29.3    &   28.6     &    20.5 & 20.4  &    21.3    &   9.6 & \textbf{29.3}    \\
    & SPA-VL           &   30.5     &   32.0     &    17.0 & 17.5  &   30.1     & 1.9  &  28.1  \\
    & BeaverTails-V    &   26.4     &   30.1     &   19.5 & 17.0  &   26.1     & 7.3  & 28.0   \\
    & MSR-Align        &   28.8     &   32.7     &   18.3 & 19.0  &   28.7     & 6.1  & \underline{28.9} \\
\midrule
\multirow{4}{*}{\shortstack[l]{InternVL2.5-1B-MPO}} 
    & Pretrained       &   33.2     &   34.3     &   20.4 & 22.0  &   29.2     & 5.8  & \textbf{32.1}    \\
    & SPA-VL           &   19.8     &   20.5     &   11.9 & 11.5  &   16.5     & 1.9  & 19.5 \\
    & BeaverTails-V    &   27.2     &   25.2     &   15.0 & 14.4  &   26.5     & 3.8  & 27.4   \\
    & MSR-Align        &   30.3     &   31.5     &   18.6 & 19.1  &   27.7     & 4.9  & \underline{30.0} \\
\midrule
\multirow{4}{*}{\shortstack[l]{VLM-R1}} 
    & Pretrained       &   39.0     &   40.8     &   26.1 & 26.6  &   33.2     & 7.7  & \textbf{38.6}   \\
    & SPA-VL           &   37.3     &   38.3     &   26.1 & 28.3  &   33.6     & 7.7  & 35.8  \\
    & BeaverTails-V    &   26.4     &   30.1     &   19.5 & 17.0  &   26.1     & 7.3  & 28.0  \\
    & MSR-Align        &   36.7     &   39.6     &   24.8 & 25.3  &   31.7     & 10.4 & \underline{36.1} \\
\bottomrule
\end{tabular}
\vskip 0.1in
\caption{MM-Vet accuracy (\%) across five reasoning domains for different models and alignment strategies. \textbf{Bold} indicates the best overall performance (total column) within each model group, and \underline{underline} marks the second-best. MSR-Align consistently ranks second while outperforming other alignment baselines.}
\end{table}

\section{Discussion}
Our findings show the crucial role of structured multimodal safety reasoning in aligning vision-language models (VLMs) for real-world deployment. Fine-tuning with {MSR-Align} consistently improves safety performance across diverse benchmarks without compromising general reasoning capabilities. This suggests that reasoning-based alignment—when grounded explicitly in visual context and policy logic—offers a viable path toward robust, high-utility multimodal alignment.

\paragraph{Structured Safety Reasoning Enables Fine-Grained Risk Detection.} Traditional alignment strategies often rely on static refusals or classification-based supervision, which tend to fall short when faced with indirect, compositional, or visually grounded unsafe instructions. MSR-Align addresses this limitation by introducing structured chain-of-thought (CoT) traces that decompose safety decisions into interpretable steps: visual grounding, rule identification, and policy-based justification. This explicit structure enables models to detect nuanced forms of unsafe content, such as when benign textual prompts are rendered harmful due to visual context, or when stepwise reasoning leads to unsafe conclusions. Our results show that this form of supervision significantly boosts model robustness in scenarios where shallow refusal signals are insufficient.

\paragraph{CoT Supervision Preserves Reasoning Utility While Enforcing Safety.} A central concern in safety alignment is the risk of over-correction—models may learn to indiscriminately refuse challenging prompts, thereby sacrificing task performance. Our evaluation on MM-Vet reveals that MSR-Align not only avoids this degradation but can even improve performance in some domains. This suggests that grounding refusal in reasoning, rather than treating it as a binary end-state, encourages models to deliberate responsibly rather than disengage entirely. By embedding policy logic within the reasoning process itself, MSR-Align equips models to explain why a task should or should not be completed, rather than simply rejecting it. This preserves the model’s utility in benign scenarios while tightening alignment in adversarial cases.

\paragraph{Multimodal Safety Demands Multimodal Reasoning Supervision.} While significant progress has been made in aligning LLMs through text-only approaches, our analysis confirms that these strategies do not generalize well to VLMs. The addition of visual input introduces compositional ambiguity and opens new attack surfaces, including misleading visual cues, deceptive spatial layouts, or plausible-looking unsafe scenarios. In these settings, final-answer refusals without intermediate supervision are brittle. MSR-Align demonstrates that aligning the full multimodal reasoning trajectory—not just the end response—is critical for ensuring reliable behavior in complex visual-language tasks. The model must learn not only what to say but how to reason about what it sees and how that affects the safety of its response.

\section{Conclusion}
This paper presents MSR-Align, a novel dataset designed to address the emerging safety challenges in reasoning-capable vision-language models (VLMs). While recent advances in chain-of-thought (CoT) reasoning have significantly improved VLM capabilities, they also introduce new vulnerabilities that existing alignment strategies—mostly developed for unimodal LLMs—fail to address. MSR-Align fills this critical gap by providing high-quality multimodal safety reasoning data grounded in standardized policy rules, enabling VLMs to deliberate more safely across both textual and visual modalities.

Our data construction framework emphasizes multimodal diversity, policy-grounded reasoning, and automatic quality filtering, resulting in a scalable pipeline for building trustworthy multimodal safety supervision. Extensive experiments across diverse open-source VLMs demonstrate that fine-tuning with MSR-Align leads to robust safety gains under both textual and vision-language jailbreak attacks, while preserving or even enhancing general reasoning performance.

Overall, this work establishes a concrete step toward safer, more trustworthy multimodal reasoning systems and highlights the importance of high-fidelity, policy-aligned reasoning data for future alignment efforts in large vision-language models.

\bibliographystyle{plainnat}  
\bibliography{neurips_2025.bib}      

\end{document}